%% file: main.tex
\documentclass{article} 
\usepackage{iclr2023_conference,times}

\input{math_commands.tex}

\usepackage{hyperref}
\usepackage{url}

\usepackage{tikz}
\usepackage{comment}
\usepackage{amsmath,amssymb} 

\usepackage{color}
\usepackage{caption}
\usepackage{subcaption}
\usepackage{xspace}
\usepackage{makecell}
\usepackage{boldline}
\usepackage{array}
\usepackage{colortbl}
\usepackage{booktabs}
\usepackage{bbm}
\usepackage{multicol}
\usepackage{multirow}
\usepackage{xspace}
\usepackage{wrapfig}

\definecolor{mygray}{gray}{.92}

\makeatletter
\DeclareRobustCommand\onedot{\futurelet\@let@token\@onedot}
\def\@onedot{\ifx\@let@token.\else.\null\fi\xspace}

\def\eg{\emph{e.g}\onedot} 
\def\ie{\emph{i.e}\onedot}

\makeatother

\title{Efficient Attention-free Video Shift Transformers}

\author{Adrian Bulat  \\
Samsung AI Center Cambridge\\
\texttt{adrian@adrianbulat.com} \\
\And
Brais Martinez \\
Samsung AI Center Cambridge\\
\texttt{brais.a@samsung.com} \\
\And
Georgios Tzimiropoulos \\
Samsung AI Center Cambridge  \\
Queen Mary University London \\
\texttt{g.tzimiropoulos@qmul.ac.uk} \\
}

\newcommand\ShortName{AST}
\newcommand\ShortNameVideo{VAST}

\iclrfinalcopy 
\begin{document}

\maketitle

\begin{abstract}
This paper tackles the problem of efficient video recognition. In this area, video transformers have recently dominated the efficiency (top-1 accuracy vs FLOPs) spectrum. At the same time, there have been some attempts \textit{in the image domain} which challenge the necessity of the self-attention operation within the transformer architecture, advocating the use of simpler approaches for token mixing. However, there are no results yet for the case of video recognition, where the self-attention operator has a significantly higher impact (compared to the case of images) on efficiency. To address this gap, in this paper, we make the following contributions: (a) we construct a highly efficient \& accurate attention-free block based on the shift operator, coined \textit{Affine-Shift} block, specifically designed to approximate as closely as possible the operations in the MHSA block of a Transformer layer. Based on our Affine-Shift block, we construct our Affine-Shift Transformer and show that it already outperforms all existing shift/MLP--based architectures for ImageNet classification. (b) We extend our formulation in the video domain to construct Video Affine-Shift Transformer (VAST), the very first purely attention-free shift-based video transformer. (c) We show that VAST significantly outperforms recent state-of-the-art transformers on the most popular action recognition benchmarks for the case of models with low computational and memory footprint. Code will be made available.
\end{abstract}

\section{Introduction}

\begin{figure}
    \centering
    \includegraphics[width=7.5cm]{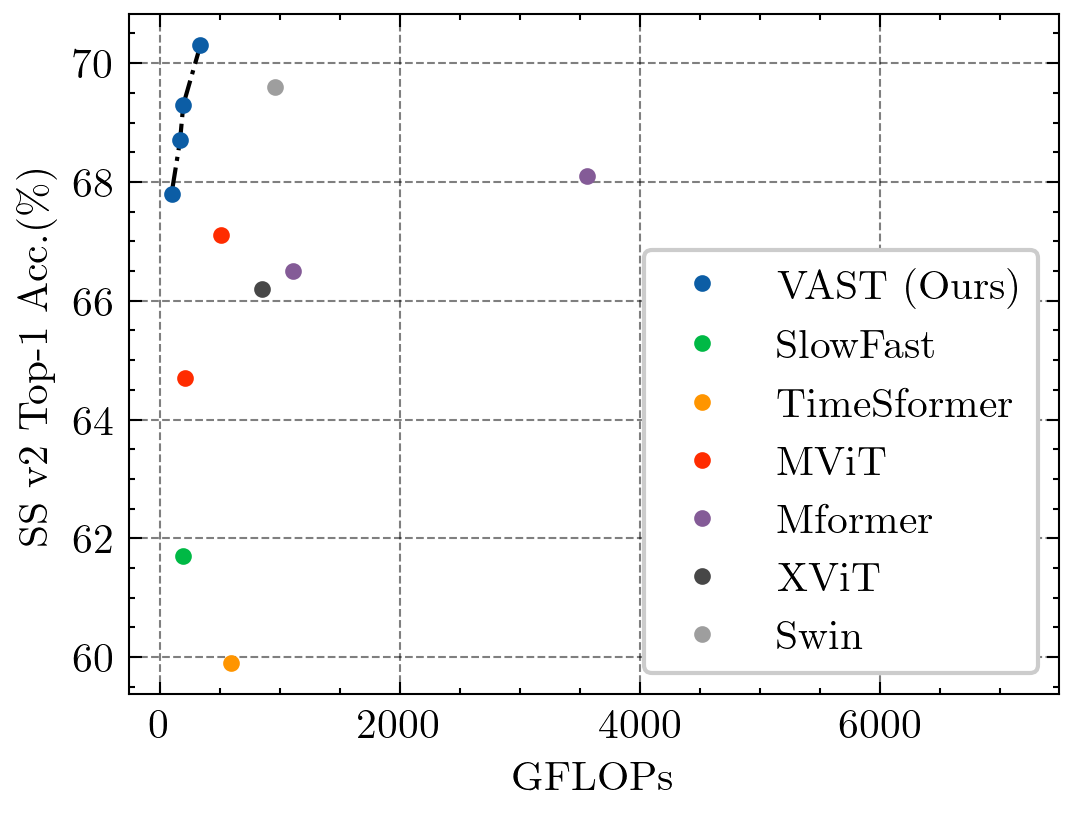}
    \caption{Our approach consistently outperforms all prior state-of-the-art in terms of efficiency-accuracy trade-off on the Something-Something-v2 dataset.}
    \label{fig:flops}
\end{figure}

Video recognition is the problem of recognizing specific events of interest (\eg actions, highlights) in video sequences. Compared to the image recognition problem, video recognition must address at least one additional important technical challenge: the incorporation of the time dimension induces significant computational overheads as, typically, in the best case, a temporal model has $T\times $ more complexity than its corresponding image counterpart ($T$ is the number of frames in the video sequence)~\footnote{This is further exacerbated by the multi-cropping inference process, a typical process in evaluating video recognition models.}. For example, existing state-of-the-art models~\cite{fan2021multiscale,bulat2021space} still require 400-1000 GFLOPs to achieve high accuracy on the Kinetics dataset~\cite{carreira2017quo}. The main result of this paper is a video recognition model than can achieve similar accuracy while \textit{requiring $\sim 3-4 \times$ less} FLOPs (see also Fig.~\ref{fig:flops}). 

Specifically, and following the tremendous success of Transformers in NLP,~\cite{vaswani2017attention,chen2018best}, the current state-of-the-art in video recognition is based on video transformers~\cite{bertasius2021space,arnab2021vivit,fan2021multiscale,bulat2021space}. While such models have achieved significantly higher accuracy compared to traditional CNN-based approaches (\eg SlowFast~\cite{feichtenhofer2019slowfast}, TSM~\cite{lin2019tsm}), they still require very large video backbones to achieve these results, \eg ViT-L/H in~\cite{arnab2021vivit}. In fact, the main reason that these models have dominated the accuracy-FLOPs spectrum is because they require significantly fewer number of crops during inference compared to CNN-based approaches.

Concurrently to the development of the aforementioned video transformers, there has been an independent line of research which questions the necessity of the self-attention layers in the vision transformer's architecture. Such ``attention-free'' methods have proposed the use of simpler schemes based on MLPs~\cite{touvron2021resmlp,chen2021cyclemlp,tolstikhin2021mlp} and/or the shift operator~\cite{yu2022s2,yu2022s1} for achieving the token mixing effect akin to the self-attention layer. However, these methods have been developed for the image domain, where the cost of the self-attention is relatively low compared to the video domain, where some approximation of the full MHSA is typically necessary.
As a result, these methods have not been conclusively shown to outperform self-attention-based transformers for image recognition. Moreover, there are no attention-free methods yet for the case of the video domain where the self-attention operation induces significantly higher computational and memory cost. Hence, the question we wish to address in this paper is: \textit{``Can we construct high performing video transformers without attention?''}

\begin{figure}
    \centering
    \includegraphics[height=6cm,trim={2.5cm 3.5cm 6.5cm 0.5cm},clip]{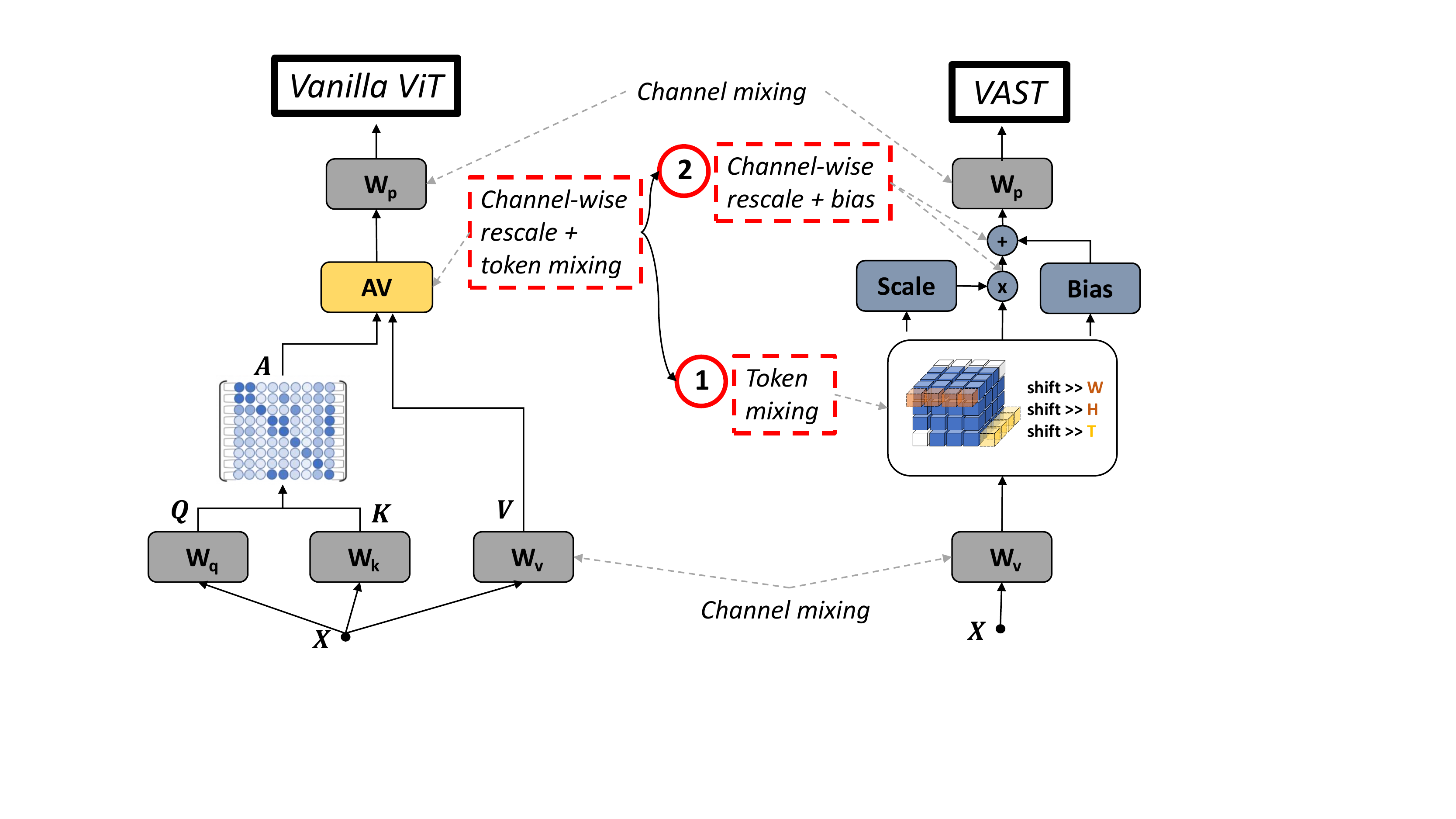}
    \caption{The proposed Affine-Shift block used to construct AST and VAST (right) and the ViT attention module (left). Our goal is to approximate as closely as possible the MHSA operations within a Transformer block: channel-wise rescaling and token mixing implemented by $\mathbf{AV}$ in MHSA (see also Eq.~\ref{eq:SA}) are now replaced by {\color{red}\textbf{(1)}} local token mixing using the shift operator followed by {\color{red}\textbf{(2)}} channel-wise rescaling (using SE-like MLP) and bias correction (using DWConv). \textit{All the above mentioned steps} are needed to obtain a highly accurate block/architecture.}

    \label{fig:overall_idea}
\end{figure}

To address the above question, \textbf{we make the following contributions}:
\begin{enumerate}
    \item We introduce a new block for attention-free transformers based on the \textit{shift operator} which is tailored to achieve high accuracy with low computational and memory footprint. Our block, coined \textit{Affine-Shift} block and shown in Fig.~\ref{fig:overall_idea}, is specifically designed to \textit{approximate} as closely as possible \textit{the operations} in the MHSA block of a Transformer layer.
    \item
    Based on our Affine-Shift block, we construct our Affine-Shift Transformer (AST). We exhaustively ablate AST in the image domain for ImageNet classification where we show that it significantly outperforms previous work particularly for the case of low complexity models.
    \item 
    By extending our Affine-Shift block in the video domain, we build a new backbone for video recognition, the proposed Video Affine-Shift Transformer (VAST). VAST has two main features: (a) it is attention-free, and (b) is purely shift-based, effectively applying, for the first time, the shift operation in \textit{both space \& time} to achieve token mixing.
    \item
    We further evaluate VAST on multiple action recognition datasets, namely Kinetics~\cite{carreira2017quo}, Something-Something-v2~\cite{goyal2017something} and Epic Kitchens~\cite{damen2018scaling} where we show that it can achieve similar accuracy to state-of-the-art video transformers, namely~\cite{fan2021multiscale} and~\cite{bulat2021space}, while \textit{requiring $\sim 3-4 \times$ less} FLOPs. 
\end{enumerate}

\section{Related work}

\noindent \textbf{Vision Transformers:} After revolutionizing NLP~\cite{vaswani2017attention,raffel2019exploring}, ViT~\cite{dosovitskiy2020image} is the first convolution-free transformer that was shown to achieve promising results on ImageNet~\cite{deng2009imagenet}. Following ViT, a number of notable extensions have been proposed~\cite{touvron2021training,liu2021swin,wang2021pyramid,chu2021twins,fan2021multiscale}. DeiT~\cite{touvron2021training} proposes a teacher-student scheme which uses a distillation token so that the student learns from the teacher through attention. PVT~\cite{wang2021pyramid} and MViT~\cite{fan2021multiscale} propose to compute the attention with a sub-sampled version of the input tokens. Swin~\cite{liu2021swin} introduces non-overlapping local windows for computing the attention and uses a shifted window operation to increase the receptive field. Twins~\cite{chu2021twins} extends PVT in a number of ways emphasizing the importance of using relative positional encodings~\cite{shaw2018self}. 

\noindent\textbf{Attention-free Transformers:} Very recently, the necessity of the self-attention operation within the ViT has been questioned by a number of works which propose spatial token mixing with MLPs~\cite{tolstikhin2021mlp,liu2021pay,touvron2021resmlp}. Moreover, such approaches have been further developed by deploying the shift operator~\cite{wu2018shift} and related variants for spatial token mixing giving rise to a number of recently proposed methods~\cite{lian2021mlp,chen2021cyclemlp,yu2022s2,yu2022s1,hou2022vision} which are more flexible by allowing the processing of images of different resolutions. AS-MLP~\cite{lian2021mlp} proposes an axial shifting strategy where features are spatially shifted in both horizontal and vertical directions. CycleMLP~\cite{chen2021cyclemlp} applies the shift operator in a cyclical fashion along the channel dimension. S$^2$-MLP~\cite{yu2022s1} groups channels together, and shifts each of these groups in a different direction. S$^2$-MLPv2~\cite{yu2022s2} extends S$^2$-MLP by expanding the channel dimension before shifting and applying a hierarchical pyramid architecture. ViP~\cite{hou2022vision} proposes to permute the height and the width dimension with the channel dimension. 

The above works have proposed attention-free architectures in the image domain where the impact of the self-attention operations on the total computational complexity is limited. Hence, the advantage of such architectures over attention-based in terms of efficiency has not been conclusively demonstrated. In \textit{our work}, we firstly propose the Affine-Shift Transformer (AST), which already outperforms all the above methods in the image domain. Then, we propose to extend it to build the Video Affine-Shift Transformer (VAST), the very first attention-free video transformer which significantly outperforms attention-based video transformers, especially for the case of low complexity and memory models.

\noindent \textbf{Video recognition:} Video recognition has been tackled over the last years using 2D+time~\cite{wang2018temporal,lin2019tsm,liu2020tam} and 3D CNN-based~\cite{tran2015learning,carreira2017quo,feichtenhofer2019slowfast} approaches. While the later are characterised by high accuracy thanks to learning strong spatio-temporal models via 3D convolutions, they yield significant computational and memory costs. Somewhere in between are 2D+time works, namely TSM and TAM~\cite{lin2019tsm,liu2020tam}, that use the shift operator~\cite{wu2018shift} for learning a temporal model at a layer level while still relying on 2D convs.

More recently, a number of video transformers ~\cite{bertasius2021space,arnab2021vivit,fan2021multiscale,bulat2021space}, extending ViT into the video domain were proposed. The main goal of these works has been primarily to reduce the cost of the full space-time attention, which is particularly memory and computationally costly, by using spatio-temporal factorization~\cite{bertasius2021space,arnab2021vivit}, low resolution self-attention and hierarchical pyramid architecture~\cite{fan2021multiscale} and space-time mixing attention~\cite{bulat2021space}. 

\noindent \textbf{Closely related works:} From the above methods, \textit{our work} is mostly related to~\cite{lin2019tsm,bulat2021space}. ~\citet{lin2019tsm}  uses the shift operator~\citep{wu2018shift} for mixing channels across time while still relying on 2D convs.~\citet{bulat2021space} proposes an efficient approximation to the full space-time attention using the shift operator. Therefore,~\citet{lin2019tsm} and~\citet{bulat2021space} rely on spatial convolutions and spatial attention, respectively, for processing information in the spatial domain. We go one step beyond and propose VAST which is, to the best of our knowledge, the first video transformer based purely on the shift operator for both space \& time processing, further showing significant computational and memory savings without compromising accuracy.

\section{Method}\label{sec:method}

\subsection{Affine-Shift \& Video Affine-Shift block}\label{ssec:affine_shift}

\noindent\textbf{Transformer block:} The basic building blocks of the Transformer~\cite{vaswani2017attention} consist of a Multi-Head Self-Attention (\texttt{MHSA}) layer followed by an \texttt{MLP} (with skip connections around them). For any transformer's layer $l$, they take the form:
\begin{eqnarray}
\label{eq:mhsa-1}
\mathbf{Y}^{l} & = & \textrm{MHSA}(\textrm{LN}(\mathbf{X}^{l-1})) + \mathbf{X}^{l-1},\\
\label{eq:mhsa-2}
\mathbf{X}^{l} & = & \textrm{MLP}(\textrm{LN}(\mathbf{Y}^{l})) + \mathbf{Y}^{l}.
\end{eqnarray}
where $\mathbf{X}^{l-1}\in\mathcal{R}^{S\times d}$ are the input features at layer $l$, $\textrm{LN}(.)$ is the Layer Norm~\cite{ba2016layer} and the  Self-Attention for a single head is given by: 

\begin{equation}
\mathbf{y}^{l}_{s} = \sum_{s'=0}^{S-1} \sigma(\frac{\mathbf{q}^{l}_{s} \cdot \mathbf{k}^{l}_{s'}}{\sqrt{d_h}}) \mathbf{v}^{l}_{s'}, \;s=0,\dots,S-1   
\label{eq:SA}
\end{equation}
where $\sigma(.)$ is the \texttt{softmax} function, $S$ is the total number of spatial locations, and $\mathbf{q}^{l}_{s},\mathbf{k}^{l}_{s}, \mathbf{v}^{l}_{s} \in\mathbb{R}^{d_h}$ are the query, key, and value
vectors computed from the input features using $\mathbf{W_q},\mathbf{W_k}, \mathbf{W_v} \in\mathbb{R}^{d \times d_h}$. The final output is obtained by concatenating and projecting the heads using $\mathbf{W_h}\in\mathbb{R}^{d \times d}$ $(d = hd_h)$.

We would like to note here that the \texttt{MLP} layers mix channel-wise information within each token, \ie act independently on each token. Instead, the mixing of information \textit{between} tokens, \ie the mixing across spatial dimensions, is exclusively carried out by the \texttt{MHSA} module. Our aim is to find an effective and efficient alternative to the \texttt{MHSA}, \ie to the token mixing component of Eq.~\ref{eq:SA}.

\noindent\textbf{Shift operator:} Our goal is to replace the \texttt{MHSA} with an attention-free alternative. One direction that showed promising results as an alternative to convolutions is channel mixing using the shift operator~\cite{wu2018shift}. The main idea is to perform data-mixing by shifting a fixed amount of channels in different directions, such that each feature vector will now contain features from adjacent locations. Note that here the term ``directions'' depends on the nature of the data (\ie for images it can mean up/down, left/right, for videos backward/forward etc.). We will denote with $\texttt{Shift}(\mathbf{X}, p, b)$ the shifting of $p$ channels from feature tensor $\mathbf{X}$ across dimensions $b\in\mathbb{N}$.

\noindent\textbf{Affine-Shift block:} Our goal is to design an attention-free transformer block, using the shift operator, which approximates as closely as possible the original transformer block. As there is no attention, there is no need to compute queries and values but we will still keep the projection matrix $\mathbf{W_v}$ to compute the values $\mathbf{V}^l$ from input features $\mathbf{X}^{l-1}$, \ie, $\mathbf{V}^l = \textrm{LN}(\mathbf{X}^{l-1})\mathbf{W_v}$. 

 A naive approach would be to replace Eq.~\ref{eq:SA} with the $\texttt{Shift}(.)$ operator, and note $b_h$ and $b_w$ as the width and height dimension indexes:
\begin{equation}
\mathbf{Z}^{l} =\texttt{Shift}(\mathbf{V}^{l}, p, [b_h,b_w]),  
\label{eq:SA-shift}
\end{equation}
Although this works, as Table~\ref{tab:attn_free_variants} shows, it is not sufficient to obtain high accuracy. While the shift operator mixes information across adjacent tokens, the signal is simply mixed but there is no scale or bias adjustment. However, as Eq.~\ref{eq:SA} shows, in self-attention, the value vectors $\mathbf{v}^{l}_{s}$ are scaled by the attention. Moreover, each channel in the output vector $\mathbf{y}^{l}_{s}$ is a linear combination of the corresponding channel of the value vectors, suggesting a channel-wise operation. 

None of these appear so far in the formulation, suggesting that an extra (channel-wise) operation is missing. To address this, we introduce an Affine-Shift operator, that uses a small \texttt{MLP} to compute a channel-wise rescaling, similar to SE-net~\cite{hu2018squeeze}, and  
a \texttt{DWConv} to compute a channel-wise bias. Notably, the scale factor and bias are computed from data in a dynamic manner (similar to the dynamic nature of the Transformer block). Moreover, both the \texttt{MLP} and the \texttt{DWConv} take as input the signal \textit{post-shifting} (as also expected from the Transformer block). The proposed Affine-Shift operation is defined as:
\begin{eqnarray}
\label{eq:shift-1}
\mathbf{Z}^{l} &=& \texttt{Shift}(\mathbf{V}^{l}, p, [b_h,b_w]),  \\
\label{eq:shift-2}
\hat{\mathbf{Z}}^{l} &=& \mathbf{Z}^l \odot \sigma(\texttt{MLP}(\texttt{AVG}(\mathbf{Z}^l))) +  \texttt{DWConv}(\mathbf{Z}^l),\label{eq:affine_shift}\\ 
\mathbf{Y}^l &=& \hat{\mathbf{Z}^l} \mathbf{W_h}  + \mathbf{X}^{l-1}
\end{eqnarray}
where $\odot$ is the Hadamard product, AVG a global average pooling layer and $\sigma$ the $\texttt{Sigmoid}$ function. Note that both the \texttt{MLP} and $3 \times 3$ \texttt{DWConv} layer introduce minimal computational overhead. Note that a final linear layer using $\mathbf{W_h}$ is applied as in the original Transformer block. 

Putting everything together, the proposed Affine-Shift block firstly applies $\mathbf{W_v}$ to obtain the values by mixing the channels, then the Affine-Shift block to mix tokens and rescale \& add bias channel-wise, and finally another projection $\mathbf{W_h}$ to mix again the channels. The block is shown in Fig.~\ref{fig:overall_idea}. Note that \textit{all the above mentioned steps} are needed to obtain a highly accurate block/architecture. See Table~\ref{tab:attn_free_variants} from Section~\ref{sec:ablations}.

\noindent\textbf{Video Affine-Shift:} For video data, we have to mix information across one extra dimension (\ie time), its index being noted as $b_t$. To accommodate this, we can naturally extend the shift operator, described in Eq.~\ref{eq:shift-1}, as follows:
\begin{equation}
    \mathbf{Z}^{l} =\texttt{Shift}(\mathbf{V}^{l}, p, [b_t, b_h, b_w]),
\end{equation}
Effectively, instead of shifting across the last two dimensions (height and width), we shift across all three: time, height, width.  Note than unless otherwise specified, the shift is applied uniformly in each (of the 3) direction. We select 1/6 channels for each direction, for a total of 1/2 channels.  Both the \texttt{MLP} and the 2D \texttt{DWConv} used to compute a dynamic scale and bias are kept as is.

\subsection{AST \& VAST architectures}\label{ssec:arch}

\begin{table*}[ht]
    \centering
\begin{tabular}{c|c|c|c|c}
    \toprule
    \renewcommand{\arraystretch}{0.1}
	 Stage & Output Size  & \ShortNameVideo-Tiny & \ShortNameVideo-Small & \ShortNameVideo-Medium \\
	\hline
	\multirow{2}{*}{I} & \multirow{2}{*}{\scalebox{1.3}{$\frac{H}{4}\times \frac{W}{4} $}} &  \multicolumn{3}{c}{$C_1=64$} \\
	\cline{3-5}
	& &  
	$\begin{bmatrix}
	\begin{array}{l}
	E_1=8 \\
	\end{array}
	\end{bmatrix} \times 3$ &
	$\begin{bmatrix}
	\begin{array}{l}
	E_1=8 \\
	\end{array}
	\end{bmatrix} \times 3$ &
	$\begin{bmatrix}
	\begin{array}{l}
	E_1=8 \\
	\end{array}
	\end{bmatrix} \times 3$ \\
	\hline
	\multirow{2}{*}{II} & \multirow{2}{*}{\scalebox{1.3}{$\frac{H}{8}\times \frac{W}{8}$}} &  \multicolumn{3}{c}{$C_2=128$} \\
	\cline{3-5}
	& & 
	$\begin{bmatrix}
	\begin{array}{l}
	E_2=8 \\
	\end{array}
	\end{bmatrix} \times 4$ &
	$\begin{bmatrix}
	\begin{array}{l}
	E_2=8 \\
	\end{array}
	\end{bmatrix} \times 4$ &
	$\begin{bmatrix}
	\begin{array}{l}
	E_2=8 \\
	\end{array}
	\end{bmatrix} \times 8$ \\ 
	\hline
	\multirow{2}{*}{III} & \multirow{2}{*}{\scalebox{1.3}{$\frac{H}{16}\times \frac{W}{16}$}}  & \multicolumn{3}{c}{$C_3=320$} \\
	\cline{3-5}
	& & 
	$\begin{bmatrix}
	\begin{array}{l}
	E_3=4 \\
	\end{array}
	\end{bmatrix} \times 8$ &
	$\begin{bmatrix}
	\begin{array}{l}
	E_3=4 \\
	\end{array}
	\end{bmatrix} \times 22$ &
	$\begin{bmatrix}
	\begin{array}{l}
	E_3=4 \\
	\end{array}
	\end{bmatrix} \times 33$ \\ 
	\hline
	\multirow{2}{*}{IV} &  \multirow{2}{*}{\scalebox{1.3}{$\frac{H}{32}\times \frac{W}{32}$}} & \multicolumn{3}{c}{$C_4\!=\!512$} \\
	\cline{3-5}
	& & 
	$\begin{bmatrix}
	\begin{array}{l}
	E_4=4 \\
	\end{array}
	\end{bmatrix} \times 3$ &
	$\begin{bmatrix}
	\begin{array}{l}
	E_4=4 \\
	\end{array}
	\end{bmatrix} \times 3$ & $\begin{bmatrix}
	\begin{array}{l}
	E_4=4 \\
	\end{array}
	\end{bmatrix} \times 3$ \\ 
\bottomrule
\end{tabular}
\vspace{0.2cm}
    \caption{Model definitions for the proposed AST and VAST. $E_i$ defines the expansion rate at each stage inside the \texttt{MLP} while the multiplier the number of blocks at the current stage. $C_i$ denotes the number of channels. $T$ is kept constant across stages.}
    \label{tab:network_configuration}
\end{table*}

Using the Affine-Shift block and its video extension, we construct the Affine-Shift Transformer (AST), and the Video Affine-Shift Transformer (VAST). We follow the standard hierarchical (pyramidal) structure for our attention-free transformers, where the resolution is dropped between stages, similar to a ResNet~\cite{he2016deep,liu2021swin,fan2021multiscale}. In all cases we use an overlapping patch embedding across space. For the time dimension the decision to enable overlaps is taken on a case-by-case basis.

For image classification (\ie ImageNet), the final predictions are obtained by taking the mean across all tokens and then feeding the obtained feature to a linear classifier. Similarly, for videos, we either form a feature representation via global pooling or aggregate the data using the temporal attention aggregation layer proposed in~\cite{arnab2021vivit,bulat2021space} before passing it to a classifier.

To differentiate between the variants of our model, we align our nomenclature to that of~\cite{dosovitskiy2020image} and detail the exact configurations in Table~\ref{tab:network_configuration}.

\section{Experimental details}\label{sec:experimental_details}

\noindent \textbf{Datasets:} We trained and evaluated our models for large-scale image recognition on ImageNet~\cite{deng2009imagenet}, and on 4 action recognition datasets, namely on Kinetics-400 and Kinetics-600~\cite{kay2017kinetics}, Something-Something-v2~\cite{goyal2017something} and Epic Kitchens-100~\cite{damen2020rescaling}. ImageNet experiments aim to confirm the effectiveness of the proposed AST compared to other recently proposed shift-based and MLP-based architectures \textit{as these works have not been applied to video domain before}. See supplementary material for a description of the datasets.

\noindent \textbf{Training details on Video:} All models, unless otherwise stated, were trained following~\cite{fan2021multiscale}: specifically, our models were trained using AdamW~\cite{loshchilov2017decoupled}  with cosine scheduler~\cite{loshchilov2016sgdr} and linear warmup for a total of 50 epochs. The base learning rate, set at a batch size of 128, was $2e-4$ ($4e-4$ for SSv2) and weight decay was $0.05$. To prevent over-fitting we made use of the following augmentation techniques: random scaling (0.08$\times$ to 1.0$\times$) and cropping, random flipping (with probability of 0.5; not for SSv2), rand augment~\cite{cubuk2020randaugment}, color jitter (0.4), mixup ($\alpha=0.8$)~\cite{zhang2017mixup} and cutmix ($\alpha=1$)~\cite{yun2019cutmix}, random erasing~\cite{zhong2020random} and label smoothing ($\lambda$ = 0.1)~\cite{szegedy2016rethinking}. During training with a 50\% probability we chose between cutmix and mixup. All augmentations are applied consistently across each frame to prevent the introduction of temporal distortions. For Kinetics we set the path dropout rate to $0.1$ while on SSv2 to $0.3$.

The models were initialised from models pre-trained on ImageNet-1k for Kinetics-400/600 and from Kinetics-400 on Something-Something-v2. When initialising from a 2D model, if a 3D patch embedding is used, we initialized it using the strategy from~\cite{fan2021multiscale}. We only use a 3D patch embedding for SSv2. The models were trained on 8 V100 GPUs using PyTorch~\cite{paszke2019pytorch}.

\noindent \textbf{Testing details on Video:} Unless otherwise stated, we used 8, 16 or 32 frames. Note that when a 3D stem is used (\ie on SSv2), the effective temporal dimension is halved. We report results for $1\times3$ views (1 temporal clip and 3 spatial crops) following~\cite{bertasius2021space,bulat2021space}.

\section{Ablation studies}\label{sec:ablations}

\subsection{Affine shift analysis and variations}\label{ssec:ablation_affine}

Firstly we analyse the impact of the three main components described in the Affine-Shift module: the shift operation, the dynamical re-scaling (\texttt{MLP}) and the bias (\texttt{DWConv}) in Eq.~\ref{eq:affine_shift}. As Table~\ref{tab:attn_free_variants} shows, replacing the \texttt{MHSA} with $\texttt{Shift}(.)$ (R1) works reasonable well and sets a strong baseline result. Adding the dynamic bias (R2) and  scale (R3) on their own improves the result in each case by almost $1.5\%$. Finally, combining the 3 components together (R4) produces the strongest result.  This showcases that all of the introduced components are necessary. 

The transformer block consists of \texttt{MHSA} and \texttt{MLP} blocks (Eq.~\ref{eq:mhsa-1}-~\ref{eq:mhsa-2}). As we already replaced the \texttt{MHSA} with $\texttt{Shift}(.)$, a natural question to ask is whether we can further improve the results by adding an additional shift within the \texttt{MLP} block in the transformer. As the results show (R5), the performance saturates and no additional gains are observed. This suggests that due to the number of layers, the effective receptive size toward the end of the network is sufficiently large to cover the entire image making additional shift operations redundant.

\begin{table}[ht!]
    \centering
    \begin{tabular}{c|c}
    \hlineB{3}
      Block variant \rule{0.5em}{0em}  & \rule{0.5em}{0em} Top-1\%  \\
    \hlineB{3}
       (R1) ours - w/o scale \& w/o bias  & 79.4  \\
       (R2) ours - w/o scale  & 80.7  \\
       (R3) ours - w/o bias & 80.9  \\
       (R4) ours & \textbf{81.8} \\
       (R5) ours + extra shift & 81.7  \\
       (R6) only shift  & 79.0  \\
    \hlineB{3}
    \end{tabular}
    \caption{Effect of various shift-based variants of our method in terms of Top-1 accuracy (\%) on ImageNet. See Section~\ref{ssec:ablation_affine} for details. All models have roughly 3.9 GFLOPs.}
    \label{tab:attn_free_variants}
\end{table}

A perhaps overlooked detail is the placement of the drop-path. Normally drop-path randomly drops the \texttt{MHSA} and/or the \texttt{MLP} block at train time. However, by-passing an Affine-Shift block will result in skipping a data mixing step, effectively producing information with slightly different spatial, temporal, or spatio-temporal context. As such for all of our model, we remove the path drop that affects our Affine-Shift block. Finally, we compare our approach with a more direct alternative, that of replacing Eq.~\ref{eq:mhsa-1} in its entirety with a shift operation. As the results show (R6), while promising, the Affine-Shift block is significantly better.

\subsection{How many channels should we shift?}

A potentially important factor that could influence the accuracy of the network is the total amount of channels shifted across all dimensions. As the results from Table~\ref{tab:num_shift_channels} show, the proposed module is generally robust to the amount of shift within the range $25\%-50\%$. We note that sustaining the accuracy at lower levels of shift is especially promising for video data, where the number of dimensions we need to shift across increases.

\begin{table}[htp!]
    \centering
    \begin{tabular}{c|cccc}
    \hlineB{3}
        & \rule{0.5em}{0em} 0\% & \rule{0.5em}{0em} 25\% &  \rule{0.5em}{0em} 33\% & \rule{0.5em}{0em} 50\%  \\
    \hlineB{3}
       \% channels shift \rule{0.5em}{0em}   & 60.2 & 81.5 & 81.8 & 81.7 \\
    \hlineB{3}
    \end{tabular}
    \vspace{0.2cm}
    \caption{Impact of the number of shifted channels on the overall accuracy in terms of Top-1 acc (\%) on ImageNet.}\label{tab:num_shift_channels}
    \vspace{-0.5cm}
\end{table}

\section{Comparison to state-of-the-art}\label{ssec:results-sota}

\begin{table*}[ht]
	\centering
    \setlength\tabcolsep{2pt}
    	\begin{tabular}{c|l|cc|cc|c}
    	    \Xhline{1.0pt}
    		\multirow{2}*{Arch.} & \multirow{2}*{Method} & \#Param & FLOPs & Train & Test & ImageNet \\
    		 ~ & ~ & (M) & (G) & Size  & Size & Top-1  \\
    	    \Xhline{1.0pt}
    		\multirow{4}{*}{\rotatebox{90}{Trans}} & DeiT-S \citep{touvron2021training} & 22 & 4.6 & 224 & 224 & 79.9 \\
    		~ & PVTv2-B2-Li \citep{Wang2021PVTv2IB} & 25 & 3.9 & 224 & 224 & 82.1 \\
    		~ & Swin-T \citep{liu2021swin} & 29 & 4.5 & 224 & 224 & 81.3 \\
    		~ & Focal-T \citep{yang2021focal} & 29 & 4.9 & 224 & 224 & 82.2 \\
    		\hline
    		\multirow{3}{*}{\rotatebox{90}{Hyb.}} & CvT-13 \citep{wu2021cvt} &  20 & 4.5 & 224 & 224 & 81.6 \\
    		~ & CoAtNet-0 \citep{dai2021coatnet} &  25 & 4.2 & 224 & 224 & 81.6 \\
    		~ & LV-ViT-S \citep{jiang2021all} & 26 & 6.6 & 224 & 224 & 83.3 \\
    		  \hline
    		  \multirow{10}{*}{\rotatebox{90}{No-attn.}} & EAMLP-14~\citep{guo2021beyond}  &    30        & $-$      &    224      &      224     &       $78.9$     \\
& ResMLP-S24~\citep{touvron2021resmlp} & 30           &  6.0      &      224      & 224          & $79.4$            \\
& gMLP-S~\citep{liu2021pay}   & 20          &  4.5     &      224      &  224         & $79.6$  \\
& GFNet-S~\citep{rao2021global} &   25	   &4.5 & 224&224 & $80.0$ \\
& GFNet-H-S~\citep{rao2021global} &  32	   &4.5 & 224&224 & $81.5$ \\
& AS-MLP-T~\citep{lian2021mlp} & 28   & 4.4 & 224 & 224 & $81.3$ \\
& CycleMLP-B2~\citep{chen2021cyclemlp} &   27&  3.9 & 224&224 & $81.6$ \\
& ViP-Small/7~\citep{hou2022vision}   & 25&  6.9 & 224&224 & $81.5$ \\
& S$^2$-MLPv2-Small/7~\citep{yu2022s2} &  25   &  6.9 & 224&224 & $82.0$     \\
& \textbf{\ShortName{}-Ti (Ours)}  & 19 & 3.9 & 224 & 224 & 81.8 \\

    	    \Xhline{1.0pt}
    	\end{tabular}
    \vspace{0.2cm}
    \caption{\textbf{Comparisons on ImageNet.} Our models are the most accurate within the ``No. attn." category. Hyb. = CNN+Transformer.}
    \label{tab:mlps}
\end{table*}  

\subsection{ImageNet-1K} 

In Table~\ref{tab:mlps}, we report results on ImageNet for the variant of most interest - tiny (\ShortName{}-Ti) of our model. Moreover, we report the results of all recently proposed Shift/MLP-- and MLP--based backbones. As it can be observed, our tiny model, \ShortName{}-Ti, is the most accurate among models of similar size with only Cycle-MLP-B2 closely following. Further results and comparison for models sizes: small (\ShortName{}-S) and medium (\ShortName{}-B) are reported in the supplementary material.

Note that our goal is not highly-accurate image recognition using very big models but efficient video recognition and hence we did not train or evaluate very big image models. The results of Table~\ref{tab:mlps} clearly show that our model is already a very good candidate for highly accurate and efficient video recognition, outperforming all other Shift/MLP-- and MLP--based approaches.

\begin{table*}[!ht]
    \centering
\resizebox{1.\textwidth}{!}{
\begin{tabular}{c|cccccc}
\hlineB{3}
Method   & Pre-train & \makecell{Top-1 \\ Acc. (\%)}  & \makecell{Top-5 \\ Acc. (\%)} & Frames & Views & \makecell{FLOPs \\ $\times10^9$}  \\ \hlineB{3}
\multicolumn{7}{c}{CNN models}  \\ \hlineB{3}
SlowFast~\citep{feichtenhofer2019slowfast}  & K-400 & 61.7 & - & $8 $ & $1\times3$ & 197 \\
TSM (R50)~\citep{lin2019tsm} & K-400 & 63.3 & 88.5 & $16 $ & $2\times 3$ & 650 \\
MSNet~\citep{kwon2020motionsqueeze} & IN-1k & 64.7 & 89.4 & $16 $ & - & - \\
TEA~\citep{li2020tea} & IN-1k & 65.1 & 89.9 & $16 $ & $10\times 3$ & 2,100 \\
bLVNet~\citep{fan2019more} & IN-1k & 65.2 & 90.3 & $32 $ & $10 \times 3$ & 3,870 \\
CT-Net~\citep{li2021ct} & IN-1k & 65.9 & 90.1 & $16 $ & $2 \times 3$ & 450 \\
TAdaConvNeXt-T~\citep{huang2022tada} & K400 & 67.1 & 90.4 & 32 & $2 \times 3$ & 564 \\
\hlineB{3}
\multicolumn{7}{c}{Transformer and Hybrid models}  \\ \hlineB{3}
TimeSformer~\citep{bertasius2021space}  & IN-21k & 59.5 & - & $96 $ & $1\times3$ & 590 \\
TimeSformer-L~\citep{bertasius2021space}  & IN-21k & 62.4 & - & $96 $ & $1\times3$ & 7,140 \\
ViViT-L/16x2~\citep{arnab2021vivit}  & IN-21k+K-400 & 65.4 & 89.8 & $32 $ & $4\times3$ & 17,352 \\
XViT-B~\citep{bulat2021space}  & IN-21k & 66.2 & 90.6 & 16 & $1\times 3$ & 850  \\
XViT-B~\citep{bulat2021space}  & K-600 & 67.2 & 90.8 & 16 & $1\times 3$ & 850  \\
MViT-B ($16\times 4$)~\citep{fan2021multiscale}  & K-400 & 64.7 & 89.2 & $16$ & $1\times 3$ & 211  \\
MViT-B ($32\times 4$)~\citep{fan2021multiscale}  & K-600 & 67.8 & 91.3 & $16$ & $1\times 3$ & 510  \\
Mformer~\citep{patrick2021keeping}  & IN-21k+K-400 & 66.5 & 90.1 & $- $ & $1\times3$ & 1,110 \\
Mformer-HR~\citep{patrick2021keeping}  & IN-21k+K-400 & 68.1 & 90.8 & $- $ & $1\times3$ & 3,555 \\
Swin-B~\citep{liu2021video}  & IN-21k+K-400 & 69.6 & 92.7 & $-$ & $1\times 3$ & 963  \\
\hlineB{3}
\multicolumn{7}{c}{MLP models (Attention-free transformers)}  \\ \hlineB{3}
\textbf{VAST-Ti (Ours)} & K-400 & 67.8 & 90.8 & 16 & $1\times 3$ & 98 \\
\textbf{VAST-Ti (Ours)} & K-400 & 69.3 & 91.3 & 32 & $1\times 3$ & 196 \\
\textbf{VAST-S (Ours)} & K-400 & 68.7 & 91.0 & 16 & $1\times 3$ & 169 \\
\textbf{VAST-S (Ours)} & K-400 & 70.9 & 92.1 & 32 & $1\times 3$ & 338 \\
\hlineB{3}
\end{tabular}
}
\vspace{0.2cm}
    \caption{Comparison with CNN-based methods and state-of-the-art video transformers on Something-Something-v2. Our tiniest model VAST-Ti-8 largely outperforms the lightest MViT (+2\%) while utilizing $2\times$ fewer FLOPs, and it is only 0.4\% behind than the lightest XViT while utilizing less than $4\times$ fewer FLOPs. Larger models show improved accuracy inducing only modest computational overheads, outperforming prior results by a large margin.}
    \label{tab:ss2}
\end{table*}

\subsection{Video Action Recognition} 

\begin{table*}[ht]
    \centering
\resizebox{1.\textwidth}{!}{
\begin{tabular}{c|cccccc}
\hlineB{3}
Method   & Pre-train & \makecell{Top-1 \\ Acc. (\%)}  & \makecell{Top-5 \\ Acc. (\%)} & Frames & Views & \makecell{FLOPs \\ $\times10^9$}  \\ \hlineB{3}
\multicolumn{7}{c}{CNN models}  \\ \hlineB{3}
TSM (R50)~\citep{lin2019tsm} & IN-1k & 74.7 & - & $16 $ & $10\times 3$ & 650 \\
SlowFast~\citep{feichtenhofer2019slowfast}  & - & 78.7 & 93.5 & $8 $ & $10\times3$ & 3,480 \\
X3D-S~\citep{feichtenhofer2020x3d}  & - & 72.9 & 90.5 & $- $ & $10\times3$ & 58 \\
X3D-XXL~\citep{feichtenhofer2020x3d}  & - & 80.4 & 94.6 & $- $ & $10\times3$ & 5,823 \\
TAdaConvNeXt-T~\citep{huang2022tada} & IN-1K & 79.1 & 93.7 & 32 & $4 \times 3$ & 1,128 \\
\hlineB{3}
\multicolumn{7}{c}{Transformer and hybrid models}  \\ \hlineB{3}
TimeSformer~\citep{bertasius2021space}  & IN-1k & 75.8 & - & $8 $ & $1\times3$ & 590 \\
TimeSformer~\citep{bertasius2021space}  & IN-21k & 78.0 & 93.7 & $8 $ & $1\times3$ & 590 \\
TimeSformer-L~\citep{bertasius2021space}  & IN-21k & 80.7 & 94.7 & $96 $ & $1\times3$ & 7,140 \\
ViViT-B/16x2~\citep{arnab2021vivit}  & IN-21k & 78.8 & - & $32 $ & $4\times3$ &3,408 \\
ViViT-L/16x2~\citep{arnab2021vivit}  & IN-21k & 80.6 & 94.7 & $32 $ & $4\times3$ & 17,352 \\
Mformer~\citep{patrick2021keeping}  & IN-21k & 79.7 & 94.2 & $- $ & $10\times3$ & 11,070 \\
Mformer-HR~\cite{patrick2021keeping}  & IN-21k & 81.1 & 95.2 & $- $ & $10\times3$ & 28,764 \\
XViT-B~\citep{bulat2021space}  & IN-21k & 78.4 & 93.7 & 8 & $1\times 3$ & 425  \\
XViT-B~\citep{bulat2021space}  & IN-21k & 80.2 & 94.7 & 16 & $1\times 3$ & 850  \\
MViT-S~\citep{fan2021multiscale}  & - & 76.0 & 92.1 & - & $1\times 5$ & 165  \\
MViT-B ($16\times 4$)~\citep{fan2021multiscale}  & - & 78.4 & 93.5 & $16$ & $1\times 5$ & 352  \\
MViT-B ($64\times 3$)~\citep{fan2021multiscale}  & - & 81.2 & 95.1 & $64$ & $3\times 3$ & 4,095  \\
En-VidTr-S~\citep{zhang2021vidtr}  & IN-21k & 79.4 & 94.0 & $8$ & $10\times 3$ & 3,900  \\
En-VidTr-M~\citep{zhang2021vidtr}  & IN-21k & 79.7 & 94.2 & $16$ & $10\times 3$ & 6,600  \\
En-VidTr-L~\citep{zhang2021vidtr}  & IN-21k & 80.5 & 94.6 & $32$ & $10\times 3$ & 11,760  \\
Swin-T~\citep{liu2021video}  & IN-1k & 78.8 & 93.6 & $-$ & $4\times 3$ & 1,056  \\
Swin-S~\citep{liu2021video}  & IN-1k & 80.6 & 94.5 & $-$ & $4\times 3$ & 1,992  \\
Swin-L (384)~\citep{liu2021video}  & IN-21k & 84.9 & 96.7 & $-$ & $10\times 5$ & 105,350  \\

\hlineB{3}
\multicolumn{7}{c}{MLP models (Attention-free transformers)}  \\ \hlineB{3}
\textbf{VAST-Ti (Ours)} & IN-1k & 78.0 & 93.2 & 8 & $1\times 3$ & 98 \\
\textbf{VAST-Ti (Ours)} & IN-1k & 79.0 & 93.8 & 16 & $1\times 3$ & 196 \\
\textbf{VAST-S (Ours)} & IN-1k & 78.9 & 93.8 & 8 & $1\times 3$ & 169 \\
\textbf{VAST-S (Ours)} & IN-1k & 80.0 & 94.5 & 16 & $1\times 3$ & 338 \\
\hlineB{3}
\end{tabular}
}
\vspace{0.2cm}
    \caption{Comparison with CNN-based methods and state-of-the-art video transformers on Kinetics-400. Our tiniest model VAST-Ti-8 largely outperforms the lightest MViT (+2\%) while utilizing $2\times$ fewer FLOPs, and it is only 0.4\% behind than the lightest XViT while utilizing less than $4\times$ fewer FLOPs. Our biggest model VAST-S-16 matches the best XViT model while utilizing less than $2\times$ fewer FLOPs.}
    \label{tab:K-400}
    \vspace{-0.5cm}
\end{table*}

We report the accuracy achieved by 4 different variants  of our models: tiny and small for both 8 and 16 frames, that is VAST-Ti-8, VAST-Ti-16 \& VAST-S-8, VAST-S-16. Our models were initialized with ImageNet-1K pre-training for K-400 and K-600, while for SSv2 we used the models trained on K-400. For SSv2, we use a 3D stem, reducing the temporal dimensionality by 2. Thus, if for example we note 32 input frames, the actual configuration (and cost) corresponds to the 16-frame VAST variants.

We compare with the state-of-the-art in video recognition: In addition to classic CNN-based approaches, we compare against early attempts in video transformers, namely TimeSformer~\cite{bertasius2021space}, ViVit~\cite{arnab2021vivit} and VidTr~\cite{zhang2021vidtr}, Mformer~\cite{patrick2021keeping}, the video version of the Swin Transformer~\cite{liu2021video} as well as the state-of-the-art, namely MViT~\cite{fan2021multiscale} and XViT~\cite{bulat2021space}. For all of these models, we have included both light and heavy versions.\\ 

\noindent\textbf{K-400 \& K-600:} Table~\ref{tab:K-400} shows our results on \textbf{K-400}. It can be seen that our tiny model VAST-Ti-8 largely outperforms all early approaches to video transformers~\citep{bertasius2021space,arnab2021vivit,patrick2021keeping,liu2021video}, as well as the lightest version of MViT (+2\%) while utilizing $2\times$ fewer FLOPs, and it is only 0.4\% behind the most efficient version of XViT (initialized in ImageNet-21K) while utilizing less than $4\times$ fewer FLOPs.  Moreover, our larger models show improved accuracy, inducing only modest computational overheads. Our biggest model VAST-S-16 matches the best XViT model while utilizing less than $2\times$ fewer FLOPs. See supplementary material for results on \textbf{K-600}, where we draw similar conclusions.

It is worth noting that XViT uses an efficient local space-time attention also based on the shift operator. Our results show that our purely shift-based model is significantly more efficient (in terms of accuracy vs FLOPs) without using any attention layers at all.\\

\noindent\textbf{SSv2:} On \textbf{SSv2}, we firstly emphasize that initialization plays a very important role and that models initialized on different dataset are hard to compare. In light of this, and since we pre-trained our models on K-400, only comparisons with methods pre-trained there (and potentially on K-600) are meaningful. As Table~\ref{tab:ss2} shows, our models significantly outperform all other models in terms of accuracy vs FLOPs. For example, our lightest model (VAST-Ti) outperforms the lightest XViT and MViT by 3.4\% and 3.1\% while 
utilizing less than $4\times$ and $2\times$ fewer FLOPs, respectively. Again, our larger models show significant accuracy improvements inducing only modest computational overheads, outperforming other models by large margin.\\
\noindent\textbf{Epic-100:} Similar conclusions can be drawn by observing our results on~\textbf{Epic-100}; see supplementary material.

\section{Conclusions}

In the paper's introduction we posed the question ``can we construct high performing video transformers without attention?'' The results provided in our results section clearly demonstrated that the answer to this question is positive. To this end, we introduced a new purely shift-based block coined \textit{Affine-Shift}, specifically designed to \textit{approximate} as closely as possible \textit{the operations} in the MHSA block of a Transformer layer. Based on our Affine-Shift block, we constructed AST and show that it outperforms previous work particularly for the case of low complexity models. By extending our Affine-Shift block in the video domain, we built VAST and then showed that it is significantly more efficient than existing state-of-the-art video transformers.

\bibliography{bibliography}
\bibliographystyle{iclr2023_conference}

\appendix
\section{Appendix}

\section{Datasets}

We trained and evaluated our models for large-scale image recognition on ImageNet~\cite{deng2009imagenet}, and on 4 action recognition datasets, namely on Kinetics-400 and Kinetics-600~\cite{kay2017kinetics}, Something-Something-v2~\cite{goyal2017something} and Epic Kitchens-100~\cite{damen2020rescaling}. ImageNet experiments aim to confirm the effectiveness of the proposed AST compared to other recently proposed shift-based and MLP-based architectures \textit{as these works have not been applied to video domain before}. 

\noindent \textbf{ImageNet:} We used the standard ImageNet-1K consisting of 1.2M training images and 50K validation images belonging to 1K classes. 

\noindent \textbf{Kinetics-400 \& 600:} The Kinetics-400 (K-400) and 600 (K-600) datasets consist of pre-segmented YouTube  clips, typically of duration of up to 10 seconds, labeled with 400 and 600 classes of human activities, respectively. As many of the original clips are no longer available, we used the ones made available by the CVD foundation \footnote{\url{https://github.com/cvdfoundation/kinetics-dataset}}. Due to the nature of the data and the actions being performed, video models strongly relying on appearance-only information already perform very well on these datasets.

\noindent \textbf{Something-Something-v2:}  The Something-Something-v2 (SSv2) dataset consists of more than 220K videos of duration between 2 and 6 seconds depicting humans performing basic actions with everyday objects. Unlike Kinetics, the dataset tends to favor models with strong temporal modeling due to the nature of the actions being performed and the fact that the objects and the backgrounds in the videos are consistent across the classes.

\noindent \textbf{Epic Kitchens-100}: The dataset is labeled using 97 verb classes and 300 noun classes. The evaluation results are reported using the standard action recognition protocol: the network predicts the ``verb'' and the ``noun'' using two heads. The predictions are then merged to construct an ``action'' which is used to report the accuracy. 

\section{Additional results on Epic Kitchens 100}

In addition to the results reported on Kinetics-400/600 and Something-Something-v2, herein we report results on the Epic Kitchens 100 dataset. As the results from Table~\ref{tab:EK-100} show, our method matches and outperforms significantly bigger models, pretrained on larger datasets.

\begin{table*}[!ht]
    \centering
\begin{tabular}{c|cccc}
\hlineB{3}
Method   & Pre-train & \makecell{Action \\ Acc. (\%)}  & \makecell{Verb \\ Acc. (\%)} & \makecell{Noun \\ Acc. (\%)}  \\ \hlineB{3}
\multicolumn{5}{c}{CNN models}  \\ \hlineB{3}
TSN~\citep{wang2018temporal} & IN-1k & 33.2 & 60.2 & 46.0 \\
TRN~\citep{zhou2018temporal} & IN-1k & 35.3 & 65.9 & 44.4 \\
TBN~\citep{kazakos2019epic} & IN-1k & 36.7 & 66.0 & 47.2 \\
TSM~\citep{lin2019tsm} & K400 & 38.3 & 67.9 & 49.0 \\
SlowFast~\citep{feichtenhofer2019slowfast}  & K400 & 38.5 & 65.6 & 50.0\\
\hlineB{3}
\multicolumn{5}{c}{Transformer models}  \\ \hlineB{3}
ViViT-L/16x2~\citep{arnab2021vivit}  & IN-21k + K400 & 44.0 & 66.4 & 56.8 \\
Mformer~\citep{patrick2021keeping}  & IN-21k + K400 & 43.1 & 66.7 & 56.5 \\
Mformer-HR~\citep{patrick2021keeping}  & IN-21k + K400 & 44.5 & 67.0 & 58.5  \\
XViT-B (x8)~\citep{bulat2021space}  & IN-21k + K400 & 41.5 & 66.7 & 53.3  \\
XViT-B (x16)~\citep{bulat2021space}  & IN-21k + K400 & 44.3 & 68.7 & \textbf{56.4} \\

\hlineB{3}
\multicolumn{5}{c}{MLP models (Attention-free transformers)}  \\ \hlineB{3}
\textbf{VAST-Ti (x8) (Ours)} & K400 & 42.3 & 69.3 & 54.0   \\
\textbf{VAST-Ti (x16) (Ours)} & K400 & \textbf{45.0} & \textbf{70.0} & 56.0  \\
\hlineB{3}
\end{tabular}
\vspace{0.2cm}
    \caption{Comparison with CNN-based methods and state-of-the-art video transformers on Epic Kitchens 100.}
    \label{tab:EK-100}
\end{table*}

\begin{table*}[tp]
	\centering
    \setlength\tabcolsep{2pt}
    	\begin{tabular}{c|l|cc|cc|c}
    	    \Xhline{1.0pt}
    		\multirow{2}*{Arch.} & \multirow{2}*{Method} & \#Param & FLOPs & Train & Test & ImageNet \\
    		 ~ & ~ & (M) & (G) & Size  & Size & Top-1  \\
    	    \Xhline{1.0pt}
    		\multirow{3}{*}{\rotatebox{90}{CNN}} & RegNetY-4G \citep{radosavovic2020designing} & 21 & 4.0 & 224 & 224 & 80.0 \\
    		~ & EfficientNet-B5 \citep{tan2019efficientnet} & 30 & 9.9 & 456 & 456 & 83.6 \\
    		~ & EfficientNetV2-S \citep{tan2021efficientnetv2} & 22 & 8.5 & 384 & 384 & 83.9 \\
    		\hline
    		\multirow{4}{*}{\rotatebox{90}{Trans}} & DeiT-S \citep{touvron2021training} & 22 & 4.6 & 224 & 224 & 79.9 \\
    		~ & PVTv2-B2-Li \citep{Wang2021PVTv2IB} & 25 & 3.9 & 224 & 224 & 82.1 \\
    		~ & Swin-T \citep{liu2021swin} & 29 & 4.5 & 224 & 224 & 81.3 \\
    		~ & Focal-T \citep{yang2021focal} & 29 & 4.9 & 224 & 224 & 82.2 \\
    		\hline
    		\multirow{3}{*}{\rotatebox{90}{Hyb.}} & CvT-13 \citep{wu2021cvt} &  20 & 4.5 & 224 & 224 & 81.6 \\
    		~ & CoAtNet-0 \citep{dai2021coatnet} &  25 & 4.2 & 224 & 224 & 81.6 \\
    		~ & LV-ViT-S \citep{jiang2021all} & 26 & 6.6 & 224 & 224 & 83.3 \\
    		  \hline
    		  \multirow{10}{*}{\rotatebox{90}{No-attn.}} & EAMLP-14~\citep{guo2021beyond}  &    30        & $-$      &    224      &      224     &       $78.9$     \\
& ResMLP-S24~\citep{touvron2021resmlp} & 30           &  6.0      &      224      & 224          & $79.4$            \\
& gMLP-S~\citep{liu2021pay}   & 20          &  4.5     &      224      &  224         & $79.6$  \\
& GFNet-S~\citep{rao2021global} &   25	   &4.5 & 224&224 & $80.0$ \\
& GFNet-H-S~\citep{rao2021global} &  32	   &4.5 & 224&224 & $81.5$ \\
& AS-MLP-T~\citep{lian2021mlp} & 28   & 4.4 & 224 & 224 & $81.3$ \\
& CycleMLP-B2~\citep{chen2021cyclemlp} &   27&  3.9 & 224&224 & $81.6$ \\
& ViP-Small/7~\citep{hou2022vision}   & 25&  6.9 & 224&224 & $81.5$ \\
& S$^2$-MLPv2-Small/7~\citep{yu2022s2} &  25   &  6.9 & 224&224 & $82.0$     \\
& \textbf{\ShortName{}-Ti (Ours)}  & 19 & 3.9 & 224 & 224 & 81.8 \\
    	    \Xhline{1.0pt}
    		\multirow{3}{*}{\rotatebox{90}{CNN}} & RegNetY-8G \citep{radosavovic2020designing} & 39 & 8.0 & 224 & 224 & 81.7 \\
    		~ & EffcientNet-B7 \citep{tan2019efficientnet} & 66 & 39.2 & 600 & 600 & 84.3 \\
    		~ & EfficientNetV2-M \citep{tan2021efficientnetv2} & 54 & 25.0 & 480 & 480 & 85.1 \\
    		\hline\
    		\multirow{3}{*}{\rotatebox{90}{Trans}} & PVT-B4 \citep{Wang2021PVTv2IB} & 62.6 & 10.1 & 224 & 224 & 83.6 \\
    		~ & Swin-S \citep{liu2021swin} & 50 & 8.7 & 224 & 224 & 83.0 \\
    		~ & Focal-S \citep{yang2021focal} & 51 & 9.1 & 224 & 224 & 83.5 \\
    		\hline
    		\multirow{3}{*}{\rotatebox{90}{Hyb.}} & CvT-21 \citep{wu2021cvt} &  32 & 7.1 & 224 & 224 & 82.5 \\
    		~ & CoAtNet-1 \citep{dai2021coatnet} &  42 & 8.4 & 224 & 224 & 83.3 \\
    		~ & LV-ViT-M \citep{jiang2021all} & 56 & 16.0 & 224 & 224 & 84.1 \\
    		\hline
    		\multirow{11}{*}{\rotatebox{90}{No. attn}} & MLP-mixer~\citep{tolstikhin2021mlp} &   59        & 11.6      &    224      &      224     &       $76.4$ \\
& EAMLP-19~\citep{guo2021beyond}  &    55        & $-$      &    224      &      224     &       $79.4$     \\
& S$^2$-MLP-deep~\citep{yu2022s1}  &  51 & 10.5 & 224 & 224 & $80.7$ \\
& CCS-MLP-36~\citep{yu2021rethinking}&   43 & 8.9 & 224 & 224 & $80.6$ \\
& GFNet-B~\citep{rao2021global} &   43	   &7.9 & 224&224 & $80.7$ \\
& GFNet-H-B~\citep{rao2021global} &    54 &	8.4 & 224&224 & $82.9$ \\
& AS-MLP-S~\citep{lian2021mlp} &   50&  8.5 & 224 & 224 & $83.1$ \\
& CycleMLP-B4~\citep{chen2021cyclemlp} &    52&  10.1 & 224&224 & $83.0$ \\
& ViP-Medium/7~\citep{hou2022vision}  & 50& 16.3  & 224&224 & $82.7$ \\
& S$^2$-MLPv2-Medium/7~\citep{yu2022s2}  &55& 16.3  & 224&224 & 83.6 \\ 
& \textbf{\ShortName{}-S (Ours)}  & 38 & 6.8 & 224 & 224 & 82.8 \\
& \textbf{\ShortName{}-B (Ours)}  & 53 & 10.2 & 224 & 224 & 83.2 \\
    	    \Xhline{1.0pt}
    	\end{tabular}
    \vspace{0.2cm}
    \caption{\textbf{Comparisons on ImageNet.} Our models are the most accurate within the ``No. attn." category. Hyb. = CNN+Transformer.}
    \label{tab:mlps-full}
\end{table*}  

\begin{table*}[!ht]
    \centering
\resizebox{1.\textwidth}{!}{
\begin{tabular}{c|cccccc}
\hlineB{3}
Method   & Pre-train & \makecell{Top-1 \\ Acc. (\%)}  & \makecell{Top-5 \\ Acc. (\%)} & Frames & Views & \makecell{FLOPs \\ $\times10^9$}  \\ \hlineB{3}
\multicolumn{7}{c}{CNN models}  \\ \hlineB{3}
LGD-3D R101 & IN-1k & 81.5 & 95.6 & $- $ & $-$ & - \\
SlowFast~\citep{feichtenhofer2019slowfast}  & - & 80.4 & 94.8 & $8 $ & $10\times3$ & 3,180 \\
SlowFast+NL~\citep{feichtenhofer2019slowfast}  & - & 81.8 & 95.1 & $16 $ & $10\times3$ & 7,020 \\
X3D-M~\citep{feichtenhofer2020x3d}  & - & 78.8 & 94.5 & $- $ & $10\times3$ & 186 \\
X3D-XL~\citep{feichtenhofer2020x3d}  & - & 81.9 & 95.5 & $- $ & $10\times3$ & 1,452 \\
\hlineB{3}
\multicolumn{7}{c}{Transformer and Hybrid models}  \\ \hlineB{3}
TimeSformer~\citep{bertasius2021space}  & IN-1k & 79.1 & 94.4 & $8 $ & $1\times3$ & 590 \\
TimeSformer-HR~\citep{bertasius2021space}  & IN-21k & 81.8 & 95.8 & $8 $ & $1\times3$ & 590 \\
TimeSformer-L~\citep{bertasius2021space}  & IN-21k & 82.2 & 95.6 & $96 $ & $1\times3$ & 7,140 \\
ViViT-L/16x2~\citep{arnab2021vivit}  & IN-21k & 82.9 & 94.6 & $32 $ & $4\times3$ & 17,352 \\
Mformer~\citep{patrick2021keeping}  & IN-21k & 81.6 & 95.6 & $- $ & $10\times3$ & 11,070 \\
Mformer-HR~\citep{patrick2021keeping}  & IN-21k & 82.7 & 65.1 & $- $ & $10\times3$ & 28,764 \\
XViT-B~\citep{bulat2021space}  & IN-21k & 82.5 & 95.4 & 8 & $1\times 3$ & 425  \\
XViT-B~\citep{bulat2021space}  & IN-21k & 84.5 & 96.3 & 16 & $1\times 3$ & 850  \\
MViT-B ($16\times 4$)~\citep{fan2021multiscale}  & - & 82.1 & 95.7 & $16$ & $1\times 5$ & 352  \\
MViT-B ($32\times 3$)~\citep{fan2021multiscale}  & - & 83.4 & 96.3 & $32$ & $1\times 5$ & 850  \\
Swin-B~\citep{liu2021video}  & IN-21k & 84.0 & 96.5 & $-$ & $4\times 3$ & 3,384  \\
Swin-L (384)~\citep{liu2021video}  & IN-21k & 86.1 & 97.3 & $-$ & $10\times 5$ & 105,350  \\

\hlineB{3}
\multicolumn{7}{c}{MLP models (Attention-free transformers)}  \\ \hlineB{3}
\textbf{VAST-Ti (Ours)} & IN-1k & 82.8 & 94.5 & 8 & $1\times 3$ & 98 \\
\textbf{VAST-S (Ours)} & IN-1k & 84.0 & 95.5 & 8 & $1\times 3$ & 169 \\
\hlineB{3}
\end{tabular}
}
\vspace{0.2cm}
    \caption{Comparison with CNN-based methods and state-of-the-art video transformers on Kinetics-600. Our tiny model VAST-Ti-8 outperforms the lightest version of MViT (+0.7\%) while utilizing $4\times$ fewer FLOPs, and even outperforms the most efficient version of XViT (+0.3\% ) while utilizing less than $4\times$ fewer FLOPs.}
    \label{tab:K-600}
\end{table*}

\begin{table*}[ht]
    \centering
\resizebox{1.\textwidth}{!}{
\begin{tabular}{c|cccccc}
\hlineB{3}
Method   & Pre-train & \makecell{Top-1 \\ Acc. (\%)}  & \makecell{Top-5 \\ Acc. (\%)} & Frames & Views & \makecell{FLOPs \\ $\times10^9$}  \\ \hlineB{3}
\multicolumn{7}{c}{CNN models}  \\ \hlineB{3}
bLVNet~\citep{fan2019more} & - & 73.4 & 91.2 & $24 \times 2 $ & $3 \times 3$ & 840 \\
STM~\citep{jiang2019stm} & IN-1k & 73.7 & 91.6 & $16 $ & - & - \\
TEA~\citep{li2020tea} & IN-1k & 76.1 & 92.5 & $16 $ & $10\times 3$ & 2,100 \\
TSM (R50)~\citep{lin2019tsm} & IN-1k & 74.7 & - & $16 $ & $10\times 3$ & 650 \\
I3D NL & IN-1k & 77.7 & 93.3 & $128 $ & $10\times 3$ & 10,800 \\
CorrNet-101 & - & 79.2 & - & $32 $ & $10\times 3$ & 6,700 \\
ip-CSN-152 & - & 79.2 & 93.3 & $8 $ & $10\times 3$ & 3,270 \\
LGD-3D R101 & - & 79.4 & 94.4 & $16 $ & $-$ & - \\
SlowFast~\citep{feichtenhofer2019slowfast}  & - & 78.7 & 93.5 & $8 $ & $10\times3$ & 3,480 \\
SlowFast~\citep{feichtenhofer2019slowfast}  & - & 79.8 & 93.9 & $16 $ & $10\times3$ & 7,020 \\
X3D-S~\citep{feichtenhofer2020x3d}  & - & 72.9 & 90.5 & $- $ & $10\times3$ & 58 \\
X3D-L~\citep{feichtenhofer2020x3d}  & - & 76.8 & 92.5 & $- $ & $10\times3$ & 551 \\
X3D-XXL~\citep{feichtenhofer2020x3d}  & - & 80.4 & 94.6 & $- $ & $10\times3$ & 5,823 \\
\hlineB{3}
\multicolumn{7}{c}{Transformer models}  \\ \hlineB{3}
TimeSformer~\citep{bertasius2021space}  & IN-1k & 75.8 & - & $8 $ & $1\times3$ & 590 \\
TimeSformer~\citep{bertasius2021space}  & IN-21k & 78.0 & 93.7 & $8 $ & $1\times3$ & 590 \\
TimeSformer-L~\citep{bertasius2021space}  & IN-21k & 80.7 & 94.7 & $96 $ & $1\times3$ & 7,140 \\
ViViT-B/16x2~\citep{arnab2021vivit}  & IN-21k & 78.8 & - & $32 $ & $4\times3$ &3,408 \\
ViViT-L/16x2~\citep{arnab2021vivit}  & IN-21k & 80.6 & 94.7 & $32 $ & $4\times3$ & 17,352 \\
Mformer~\cite{patrick2021keeping}  & IN-21k & 79.7 & 94.2 & $- $ & $10\times3$ & 11,070 \\
Mformer-HR~\citep{patrick2021keeping}  & IN-21k & 81.1 & 95.2 & $- $ & $10\times3$ & 28,764 \\
XViT-B~\citep{bulat2021space}  & IN-21k & 78.4 & 93.7 & 8 & $1\times 3$ & 425  \\
XViT-B~\citep{bulat2021space}  & IN-21k & 80.2 & 94.7 & 16 & $1\times 3$ & 850  \\
MViT-S~\citep{fan2021multiscale}  & - & 76.0 & 92.1 & - & $1\times 5$ & 165  \\
MViT-B ($16\times 4$)~\citep{fan2021multiscale}  & - & 78.4 & 93.5 & $16$ & $1\times 5$ & 352  \\
MViT-B ($64\times 3$)~\citep{fan2021multiscale}  & - & 81.2 & 95.1 & $64$ & $3\times 3$ & 4,095  \\
En-VidTr-S~\citep{zhang2021vidtr}  & IN-21k & 79.4 & 94.0 & $8$ & $10\times 3$ & 3,900  \\
En-VidTr-M~\citep{zhang2021vidtr}  & IN-21k & 79.7 & 94.2 & $16$ & $10\times 3$ & 6,600  \\
En-VidTr-L~\citep{zhang2021vidtr}  & IN-21k & 80.5 & 94.6 & $32$ & $10\times 3$ & 11,760  \\
Swin-T~\citep{liu2021video}  & IN-1k & 78.8 & 93.6 & $-$ & $4\times 3$ & 1,056  \\
Swin-S~\citep{liu2021video}  & IN-1k & 80.6 & 94.5 & $-$ & $4\times 3$ & 1,992  \\
Swin-B~\citep{liu2021video}  & IN-1k & 80.6 & 94.6 & $-$ & $4\times 3$ & 3,384  \\
Swin-L (384)~\citep{liu2021video}  & IN-21k & 84.9 & 96.7 & $-$ & $10\times 5$ & 105,350  \\

\hlineB{3}
\multicolumn{7}{c}{MLP models (Attention-free transformers)}  \\ \hlineB{3}
\textbf{VAST-Ti (Ours)} & IN-1k & 78.0 & 93.2 & 8 & $1\times 3$ & 98 \\
\textbf{VAST-Ti (Ours)} & IN-1k & 79.0 & 93.8 & 16 & $1\times 3$ & 196 \\
\textbf{VAST-S (Ours)} & IN-1k & 78.9 & 93.8 & 8 & $1\times 3$ & 169 \\
\textbf{VAST-S (Ours)} & IN-1k & 80.0 & 94.5 & 16 & $1\times 3$ & 338 \\
\hlineB{3}
\end{tabular}
}
\vspace{0.2cm}
    \caption{Comparison with CNN-based methods and state-of-the-art video transformers on Kinetics-400. Our tiniest model VAST-Ti-8 largely outperforms the lightest MViT (+2\%) while utilizing $2\times$ fewer FLOPs, and it is only 0.4\% behind than the lightest XViT while utilizing less than $4\times$ fewer FLOPs. Our biggest model VAST-S-16 matches the best XViT model while utilizing less than $2\times$ fewer FLOPs.}
    \label{tab:K-400-full}
\end{table*}

\end{document}

%% file: math_commands.tex
\usepackage{amsmath,amsfonts,bm}

\def\eqref#1{equation~\ref{#1}}

\def\1{\bm{1}}

\DeclareMathAlphabet{\mathsfit}{\encodingdefault}{\sfdefault}{m}{sl}
\SetMathAlphabet{\mathsfit}{bold}{\encodingdefault}{\sfdefault}{bx}{n}

